%% file: main.tex
\documentclass[10pt,twocolumn,letterpaper]{article}
\usepackage[T1]{fontenc}
\usepackage[pagenumbers]{cvpr}
\input{preamble}
\input{paper_config}
\title{\paperTitle}
\author{\paperAuthors}
\hypersetup{pdftitle={\paperTitle},pdfauthor={\paperAuthorMetadata}}
\usepackage{placeins}
\begin{document}
  \twocolumn[{%
    \renewcommand{\twocolumn}[1][]{##1}%
    \maketitle
    \ifdefempty{\paperAuthors}{\vspace{-24pt}}{}%
    \begin{minipage}{\textwidth}
      \input{sections/00_teaser}
    \end{minipage}
    \vspace{1em}%
  }]
\input{sections/00_abstract}
\input{sections/01_introduction}
\input{sections/02_related_work}
\input{sections/03_method}
\input{sections/03_dataset}
\input{sections/04_experiments}
\input{sections/05_conclusion}
{\small
\bibliographystyle{ieeenat_fullname}
\bibliography{references}
}
\clearpage
\appendix
\raggedbottom
\clearpage
\appendix

% CVPR 模板提供的补充材料标题
% \maketitlesupplementary

% 补充材料内容
% \input{citylift_supplementary}

\end{document}

%% file: preamble.tex
\usepackage{graphicx,amsmath,amssymb,booktabs,array,tikz}
\usepackage{xcolor,xspace,microtype,enumitem,xurl}
\usepackage{colortbl,etoolbox,float}

\usepackage{hyperref}
\hypersetup{hidelinks}
\usepackage{multirow}
\ifdefined\pdfminorversion
\fi

\definecolor{CLours}{HTML}{EFF5FB}
\definecolor{CLbestbg}{HTML}{DCEFE7}
\definecolor{CLbestfg}{HTML}{155C4E}
\definecolor{CLframe}{HTML}{CDD7E2}
\definecolor{CLplaceholder}{HTML}{F7F9FC}
\definecolor{CLlabel}{HTML}{506277}

\providecommand{\CLfigure}[4]{%
  \IfFileExists{#1}{%
    \includegraphics[width=\linewidth]{#1}%
  }{%
    \begingroup
    \setlength{\fboxsep}{7pt}%
    \fcolorbox{CLframe}{CLplaceholder}{%
      \begin{minipage}[c][#2][c]{%
        \dimexpr\linewidth-2\fboxsep-2\fboxrule\relax}
        \centering\color{CLlabel}\sffamily
        {\small\bfseries #3}\par\medskip
        {\footnotesize Figure space reserved}\par\smallskip
        {\footnotesize #4}\par
      \end{minipage}%
    }%
    \endgroup
  }%
}

%% file: paper_config.tex
\newcommand{\paperTitle}{CityLift: Compositional 3D Urban Generation from a Single Image}

\newcommand{\paperAuthors}{
  Haoran Gong$^{1,2}$ \quad Jie Feng$^{2,*}$ \quad Chenxu Wang$^{1,*}$\\
  $^{1}$School of Software Engineering, Xi'an Jiaotong University\\
  $^{2}$Zhongguancun Academy, Beijing, China\\
  {\small $^{*}$Corresponding author.}
}
\newcommand{\paperAuthorMetadata}{Haoran Gong, Jie Feng, Chenxu Wang}

%% file: sections/00_teaser.tex
% Full-width teaser, inserted together with the title by main.tex.
\centering
\captionsetup{hypcap=false}
\includegraphics[width=\linewidth]{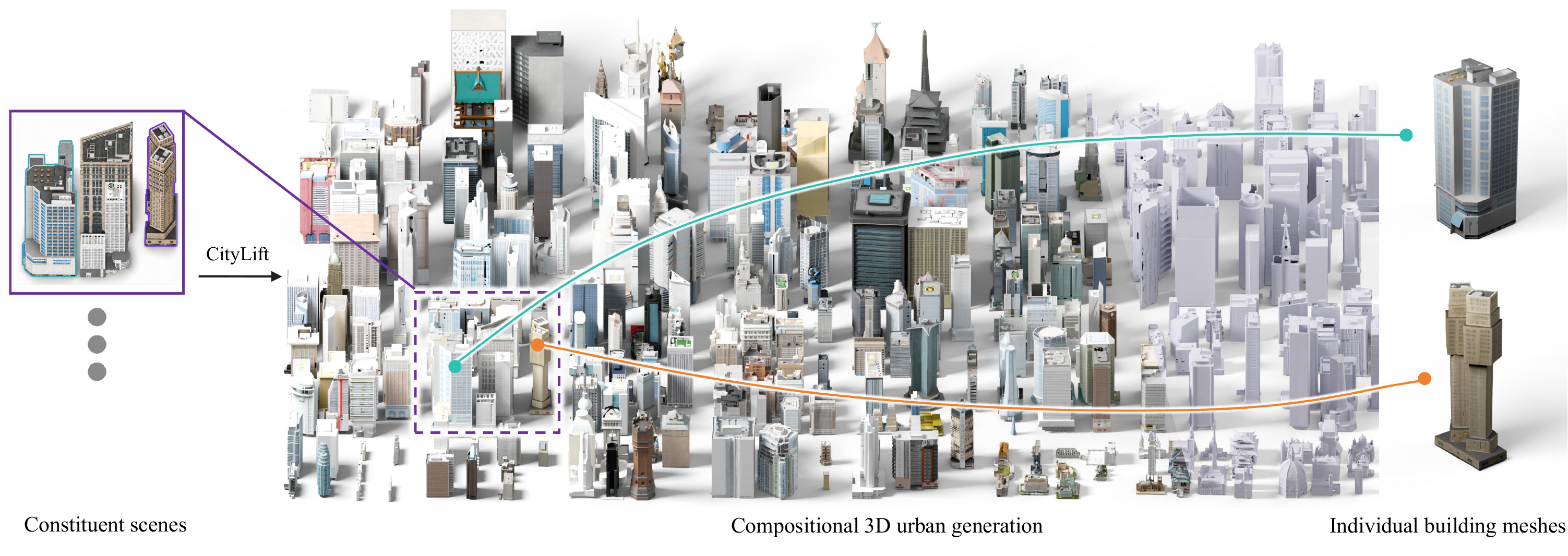}
\captionof{figure}{Overview of \textbf{CityLift.} Given a single urban image and its building instance masks, CityLift generates independently accessible textured building meshes while preserving their spatial composition.}
\label{fig:teaser}

%% file: sections/00_abstract.tex
\begin{abstract}

Generating editable 3D urban scenes from a single image requires detailed building assets with a coherent spatial arrangement. Occlusion and perspective ambiguity make it difficult to infer complete building geometry together with consistent relative scales and positions. We present CityLift, a framework that generates individual textured building meshes from an urban image by coupling scene-level spatial modeling with building-level synthesis. CityLift first learns building shape and placement jointly through coupled scene and instance streams, producing coarse sparse structures in a common coordinate system. We then introduce generative spatial inheritance, which maps each building's coarse occupancy and scene-conditioned 3D feature field into a local volume to guide a flow model that generates finer structure. The same coordinate mapping places the generated meshes back into the scene, preserving their spatial relationship to the scaffold. To address the scarcity of urban data pairing scene images with individual 3D buildings, we construct CityLift-5K, comprising 4,997 synthetic scenes with instance masks, building meshes, and placement annotations. Experiments on CityLift-5K and scene blocks from UrbanScene3D and InstanceBuilding demonstrate improved building
geometry and scene composition. The project page is available at
\url{https://ranhaogong.github.io/CityLift/}.
\end{abstract}

%% file: sections/01_introduction.tex
\section{Introduction}
\label{sec:introduction}
3D urban generation supports embodied intelligence, urban
simulation, and urban world modeling~\citep{shang2024urbanworld, liu2026spatialact, feng2025urbanllava}.
Existing approaches employ neural scene
representations~\citep{xie2024citydreamer,xie2025gaussiancity},
asset retrieval and language-guided
planning~\citep{deng2024citycraft,shang2024urbanworld,huang2025majutsucity},
or satellite-conditioned mesh
generation~\citep{hua2025sat2city,hua2026sat2cityv2}.
Neural representations do not directly yield complete building
meshes, while unified city meshes require further decomposition
to obtain independently editable assets.
Retrieval constrains geometric diversity to the available
library, and iterative planning and refinement incur additional
inference costs.

Scalable data engines for urban world models and embodied
intelligence require generators that produce explicit,
textured building assets, support instance-level manipulation,
and generalize to unseen buildings and layouts with efficient
inference~\citep{shang2024urbanworld, wang2026urbanworld2}. 
These requirements motivate us to investigate
\emph{single-image compositional urban generation}, using an
oblique urban image as a joint specification of architectural
appearance and spatial composition.
Given this image and building instance masks, the task generates
a complete textured mesh for each building in a shared scene
coordinate system.
The resulting assets remain independently editable, with
relative positions, orientations, scales, and heights consistent
with the input image.

Compositional image-to-scene methods~\citep{huang2025midi,meng2026scenegen,yin2026fixer}
demonstrate the potential of generating multiple assets from one image. In urban blocks, however, three key challenges arise.
(1) \emph{Scene consistency.} Generated buildings must maintain
coherent relative positions, orientations, scales, and heights.
Occlusion and height variation make these quantities ambiguous,
leaving scene assembly sensitive to errors in explicit pose
prediction~\citep{sam3dteam2025sam3d} or estimated geometric
anchors~\citep{yin2026fixer,sun2026mirascene}.
(2) \emph{Geometric fidelity across scales.} Joint generation
in a shared scene frame models geometry and placement
together~\citep{ling2025iscene}, but a fixed-resolution grid
covering an entire block restricts the geometric detail
available to individual buildings~\citep{zang2026hive3d}.
Increasing instance resolution therefore requires preserving
spatial correspondence between local geometry and the scene.
(3) \emph{Limited urban supervision.} Established scene
datasets emphasize indoor compositions~\citep{fu20213dfront},
while recent urban resources provide individual building
assets~\citep{jin2026satskylines,wang2025sat2realcity}
or imagery paired with regional scene
meshes~\citep{hua2026sat2cityv2}.
Learning urban composition and building completion jointly
requires supervision that combines these levels, pairing
oblique block images with complete building instances and
their spatial arrangement.

To address these challenges, we introduce CityLift, a generative
framework that models urban composition in a shared scene space
and synthesizes detailed buildings in dedicated local spaces.
This design reconciles the need for joint spatial reasoning
across buildings with the resolution required for individual
architectural detail. A jointly generated urban scaffold connects
these scales, providing the spatial reference and contextual
conditions for local building generation.

Specifically, \emph{Urban Scaffold Generation} jointly generates
coarse building geometry and layout through coupled scene and
instance streams, encoding placement directly in scene-space
geometry without a separate pose-regression stage.
To retain this spatial organization during local generation,
\emph{Generative Spatial Inheritance} maps scaffold occupancy
and scene-conditioned features into aligned building volumes.
Within these volumes, \emph{Conditioned Building Regeneration}
synthesizes finer sparse structures for textured mesh generation.
A shared coordinate mapping governs local conditioning,
supervision, and assembly, concentrating geometric resolution
on individual buildings while maintaining their correspondence
to the scene scaffold.

Training this framework requires paired supervision for scene
composition and complete building geometry.
We therefore curate building assets for geometric completeness
and texture quality through vision-language-model-assisted
filtering and manual screening, and construct \emph{CityLift-5K}.
The dataset comprises 4,997 synthetic multi-building scenes
with paired oblique images, instance masks, complete building
meshes, and placement annotations, providing supervision
at both scene and building levels.

Trained on CityLift-5K, CityLift achieves state-of-the-art
geometric performance on its held-out test split and two
external datasets.
On the CityLift-5K test split, it reduces building- and
scene-level Chamfer distances by 60.3\% and 58.4\%, respectively,
relative to the strongest baseline at each level.
Further evaluation on synthetic scenes from UrbanScene3D
and real-world scenes from InstanceBuilding yields the best
scene-level geometric results on both datasets, demonstrating
strong generalization capability.

Our main contributions are as follows.
\begin{itemize}[leftmargin=*,itemsep=3pt,topsep=3pt]
\item We present CityLift, a framework for compositional urban
generation from a single oblique image and building instance
masks.
Its \emph{Urban Scaffold Generation} jointly models coarse
building geometry and layout, establishing shared spatial
support for the synthesis and assembly of complete,
independently editable textured building meshes.

\item We introduce \emph{Generative Spatial Inheritance}
to align coarse occupancy and scene-conditioned hidden
features with building-specific local volumes, and
\emph{Conditioned Building Regeneration} to synthesize
finer sparse structures under these inherited conditions.

\item We construct \emph{CityLift-5K}, a dataset of 4,997
synthetic urban scenes. CityLift achieves state-of-the-art
scene-level geometric performance on CityLift-5K,
UrbanScene3D, and real-world scenes from InstanceBuilding.
\end{itemize}

% \begin{itemize}[leftmargin=*,itemsep=3pt,topsep=3pt]
% \item We introduce CityLift, a framework for generating urban blocks composed of individual textured building meshes from a single image and instance masks. Its \emph{Urban Scaffold Generation} jointly infers building structure and layout, establishing a shared spatial reference for subsequent instance generation.
% \item We propose \emph{Generative Spatial Inheritance} and \emph{Conditioned Building Regeneration} to transfer coarse occupancy and scene-conditioned hidden features into detailed building synthesis. These modules overcome the limited per-building resolution of the scene scaffold through local generation, while maintaining coordinate correspondence across conditioning, supervision, and scene assembly.
% \item We construct \emph{CityLift-5K} to address the scarcity of paired urban scene and building data. The dataset comprises 4,997 synthetic multi-building scenes, pairing images and instance masks with complete building meshes and their spatial placements to support both layout learning and detailed geometry generation.
% \end{itemize}

%% file: sections/02_related_work.tex
\section{Related Work}
\label{sec:related}

% Generating an editable urban block from a single image requires both complete building assets and coherent spatial arrangement. We therefore review urban generation, single-object 3D generation, and 3D scene generation to position our task and framework within existing approaches to asset synthesis and scene composition.

\subsection{3D urban generation}
Urban generation methods differ in how they specify city structure and construct its constituent buildings.
Program-generating
methods~\citep{zhang2024cityx,liu2026citygenagent}
construct editable cities through rules and executable programs,
with architectural variation governed by the available
procedural operations.
Layout-conditioned
generators~\citep{lin2023infinicity,xie2024citydreamer,
xie2025gaussiancity,lu2024urbanarchitect}
synthesize urban appearance from spatial controls, but their
neural representations do not directly provide complete building
meshes for independent editing and reuse.
Asset retrieval~\citep{deng2024citycraft} provides explicit
building meshes, with geometric diversity constrained by the
asset library.
Generative asset
pipelines~\citep{shang2024urbanworld,huang2025majutsucity,
wang2026urbanworld2}
support editable scene construction through map- or
language-guided planning, requiring scene structure to be
established through layouts or geospatial information. Satellite-conditioned methods introduce image-based scene
control. Sat2City and its
extension~\citep{hua2025sat2city,hua2026sat2cityv2}
generate scene-level meshes, which require additional instance
decomposition for building-level manipulation.
CitySculpt~\citep{yao2025citysculpt} refines building geometry
initialized from estimated height maps, while
Sat2RealCity~\citep{wang2025sat2realcity}
uses OSM constraints to spatially organize generated buildings.
These building-level approaches derive scene structure from
overhead geometric estimates or geographic priors.

CityLift introduces \emph{single-image compositional urban
generation}, jointly generating complete textured building
meshes and their spatial arrangement from an oblique urban
image and building instance masks. This task combines
image-based control over building appearance and block
composition with independently editable assets, enabling
buildings to be replaced, rearranged, and reused in
downstream environments.

\subsection{Single-image 3D object generation}
Object generators provide priors for completing geometry beyond the visible surface, with different mechanisms for incorporating image evidence.
Optimization-based methods distill image diffusion priors into a 3D representation~\citep{tang2024dreamgaussian,sun2024dreamcraft3d}, while view-synthesis approaches generate additional observations to constrain reconstruction~\citep{liu2023zero123,shi2023zero123plus,liu2024syncdreamer,long2024wonder3d}.
Feed-forward reconstruction models amortize this inference, mapping an input image or synthesized views to meshes, radiance fields, or Gaussians~\citep{hong2024lrm,tochilkin2024triposr,tang2024lgm,wang2024crm,xu2024instantmesh}.
Native 3D generators model shape distributions directly in learned latent spaces~\citep{zhang2024clay,wu2024direct3d,li2025triposg,zhao2025hunyuan3d2}; TRELLIS and TRELLIS.2 organize geometry and appearance through structured spatial latents~\citep{xiang2025trellis,xiang2026trellis2}.
Beyond isolated assets, SAM~3D generates masked objects from natural images while also estimating layout variables~\citep{sam3dteam2025sam3d}. Through large-scale pretraining, these models learn rich shape and appearance priors that provide a foundation for building synthesis with limited paired urban data. CityLift draws on these priors through the SLAT generation and mesh decoding stages of SAM 3D Objects for textured building synthesis.

\subsection{3D scene generation}
Scene generation requires coupling object geometry with a shared spatial arrangement.
Indoor layout generators predict object attributes or plan asset placement before assembling a scene~\citep{paschalidou2021atiss,tang2024diffuscene,yang2024holodeck}.
For single-image input, compositional generators couple objects or parts during denoising~\citep{huang2025midi,lin2025partcrafter}, while other architectures represent spatial information through explicit position prediction~\citep{meng2026scenegen} or pixel-aligned 3D feature volumes~\citep{li2026pixal3d}.
An alternative separates asset synthesis from scene assembly, using geometric correspondences or layout optimization to establish object placement~\citep{ardelean2025gen3dsr,han2025reparo,yao2025cast,sun2026mirascene}.
Mira-Scene aligns canonical object coordinates with a predicted scene point cloud to recover object transformations.
Structure-conditioned refinement places greater emphasis on coarse geometric support: HIVE-3D extracts and hierarchically refines components from a generated scene~\citep{zang2026hive3d}, whereas 3D-Fixer completes objects in place using estimated visible geometry, without separate pose alignment~\citep{yin2026fixer}.
% Scene expansion offers another route to scale, extending object generators through contextual tiles or overlapping spatial windows~\citep{engstler2025syncity,engstler2026syncity3000,yoon2026extend3d}.
CityLift shares the coarse-to-fine organization of component refinement, with a focus on the coupled placement, height, and geometric detail of urban buildings.
Its link between stages is the joint transport of generated occupancy and hidden 3D features into building-specific volumes.
Building arrangement is learned in scene coordinates, while local regeneration allocates resolution to each instance.

%% file: sections/03_method.tex
\section{Method}

\begin{figure*}[t]
    \centering
    \includegraphics[width=\textwidth]{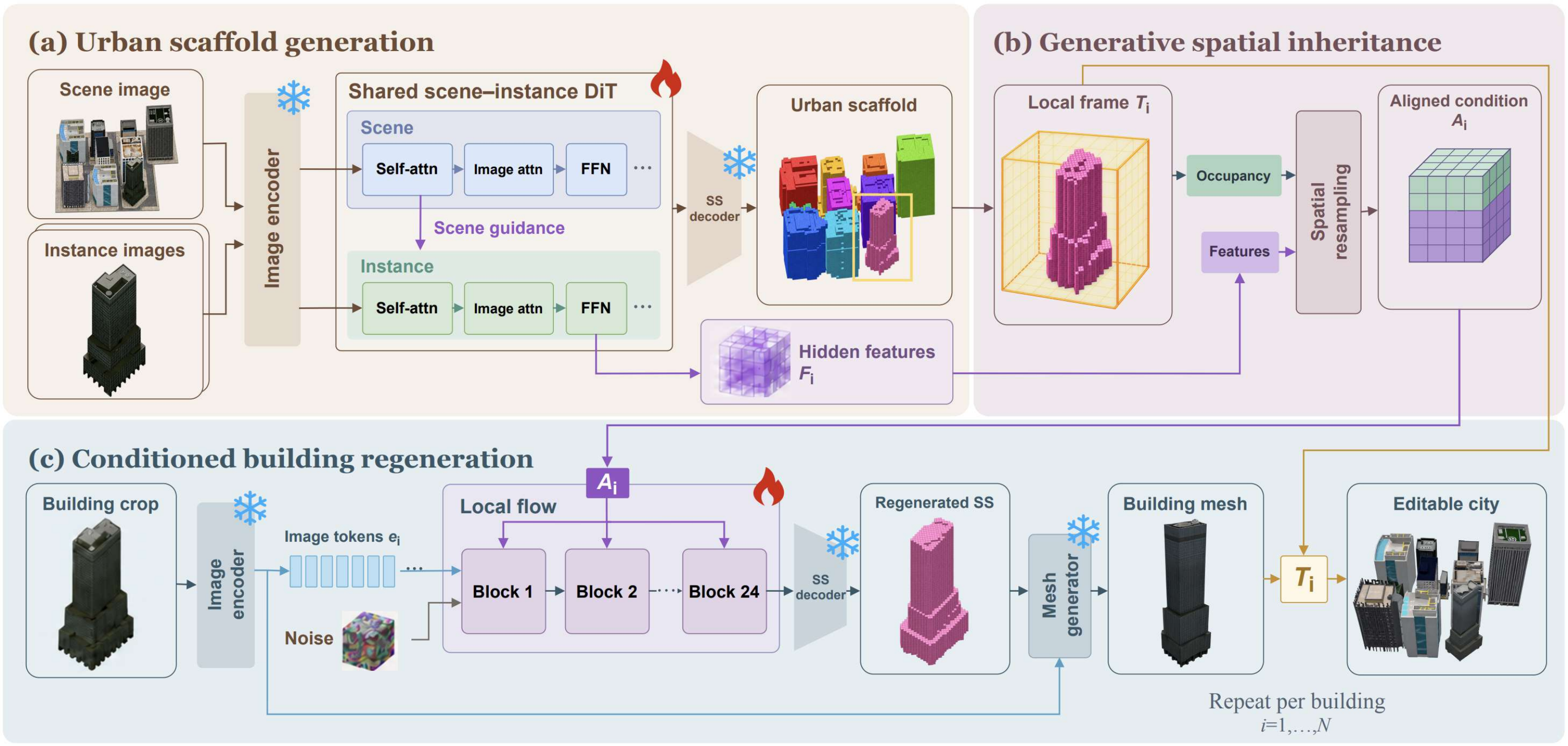}
    \caption{
        \textbf{Overview of CityLift.}
        Given a single urban image and building instance masks,
        CityLift generates individually editable building meshes
        in a shared scene frame.
        \textbf{(a)} A shared scene--instance generator predicts coarse
        occupancy in a common scene frame, implicitly modeling building
        placement, scale, and geometry.
        \textbf{(b)} A scaffold-derived coordinate map $T_i$ deterministically
        resamples occupancy and contextual features $F_i$ into each
        building's local volume, forming the aligned condition $A_i$.
        \textbf{(c)} Conditioned on $A_i$ and building image features $e_i$,
        a local flow model regenerates finer sparse structures from noise.
        These structures guide mesh generation, and the resulting buildings
        are mapped back into the scene using the inherited transforms.
    }
    \label{fig:method}
\end{figure*}

\label{sec:method}

\subsection{Problem Definition}
Given a single oblique urban image $I$ and building instance masks
$\{M_i\}_{i=1}^{N}$, our goal is to generate a scene
$\mathcal S=\{\mathcal M_i\}_{i=1}^{N}$ comprising textured building
meshes in a common normalized domain
$\Omega_s=[-\tfrac12,\tfrac12]^3$.
Each mesh $\mathcal M_i$ corresponds to mask $M_i$ and remains
independently editable.
The task requires generating plausible building geometry and appearance
while inferring placement, orientation, and relative dimensions,
including height and footprint size, consistent with the input image.

\subsection{Overview of CityLift Pipeline}
\label{sec:overview}
Our design separates the spatial extent allocated to an instance from the geometry synthesized within it. The scene generator establishes coarse support $C_i$ and a hidden feature field $F_i$ under image context. These define a \emph{generative scaffold}: an explicit geometric summary paired with a learned representation indexed by the same scene coordinates. A building-specific map $T_i$ transports the scaffold into a local volume, where a conditional flow generates refined structure $\widehat Y_i$:
\begin{equation}
 \begin{gathered}
 (C_i,F_i)\sim\mathcal G_{\phi}(I,M_i,M_s),\\
 A_i=\mathcal P_{T_i}(C_i,F_i),\qquad
 \widehat Y_i\sim p_{\theta}(Y\mid e_i,A_i).
 \end{gathered}
 \label{eq:overview}
\end{equation}
Here $M_s$ is the union of building masks, $e_i$ contains the local building-image tokens and $\mathcal P_{T_i}$ is the spatial transport operator. The map $T_i$ is computed from the predicted support. It is reused to construct the local training target and to assemble the generated mesh, connecting all three operations to the same scene-space reference. Figure~\ref{fig:method} summarizes this interface.

\subsection{Urban Scaffold Generation}
\label{sec:scene_structure}

Local building generation requires a spatial reference that specifies
each instance's location and extent within the block.
We obtain this reference by generating building occupancy directly in
$\Omega_s$, where shape, placement, and relative scale are encoded
by the same occupied cells.
To infer these properties from partially occluded observations,
the generator couples the target building's appearance with an evolving
representation of the surrounding building ensemble.

\paragraph{Scene guidance.}
A scene stream and an instance stream share the parameters of a
sparse-structure DiT.
Their image conditions are obtained from $I\odot M_s$ and
$I\odot M_i$, respectively, where $M_s=\bigvee_{j=1}^{N}M_j$.
Both conditions use the same crop and resize, preserving the target's
projected position and size within the block.
A frozen image encoder maps them to tokens $e_s^{g}$ and $e_i^{g}$.

The two streams process independently noised SS latents at a common
flow time.
At selected transformer blocks, scene guidance extends the instance
stream's attention to the scene representation.
For each attention head, the instance update is
\begin{equation}
\begin{aligned}
L_{ir} &= Q_i K_r^{\mathsf T}/\sqrt{d_h},
\qquad r\in\{i,s\},\\
U_i &= \operatorname{softmax}\!\left([L_{ii},L_{is}]\right)
\begin{bmatrix}V_i\\V_s\end{bmatrix},
\end{aligned}
\label{eq:scene_guidance}
\end{equation}
where $Q_i$, $K_r$, and $V_r$ are the projected attention tensors
and $d_h$ is the head dimension.
This joint normalization lets instance queries aggregate information
from both their own structure and the scene.
The scene stream attends within itself.
Each stream then applies image cross-attention with its corresponding
image tokens and a feed-forward layer.
Other blocks retain within-stream self-attention.

\paragraph{Learning the scaffold.}
We jointly train both streams with flow matching using
scene-coordinate occupancy targets.
The instance target $C_i^{\star}$ retains the building's placement,
while $C_s^{\star}=\bigvee_{j=1}^{N}C_j^{\star}$ represents the ensemble.
Their shared normalization makes spatial arrangement part of the
geometry supervision.
At inference, the SS decoder produces coarse occupancies $C_i$ and
$C_s$ at resolution $R_s=64$.
We retain $C_i$ together with the instance stream's hidden 3D feature
field $F_i$ before the output projection.
This pair forms the urban scaffold.
The occupancy determines the building's spatial support, while the
features retain scene-conditioned information at corresponding
3D locations for subsequent spatial inheritance.

\subsection{Generative Spatial Inheritance}
\label{sec:spatial_transfer}

The scene grid distributes its resolution across multiple buildings.
To allocate greater geometric capacity to each instance, we transfer
its scaffold into a dedicated local volume.
A deterministic coordinate map preserves the correspondence with
$\Omega_s$, allowing local generation to retain the spatial reference
established by the scaffold.

\paragraph{Inherited local frame.}
For each nonempty $C_i$, we construct a padded, axis-aligned cube
enclosing its occupied cells.
Let $\mathbf c_i$ be its center in scene coordinates and $s_i$ its
side length measured in scene-grid cells.
The local domain $\Omega_i=[-\tfrac12,\tfrac12]^3$ maps to the scene as
\begin{equation}
T_i(\mathbf u)=\mathbf c_i+\alpha_i\mathbf u,
\qquad \alpha_i=\frac{s_i}{R_s}.
\label{eq:transform}
\end{equation}
This map preserves the scene axes.
An $R_l^3$ local grid, with $R_l=64$, assigns each building a dedicated
voxel budget while retaining its scale through $\alpha_i$.

\paragraph{Aligned scaffold transfer.}
We use $T_i$ to sample occupancy and hidden features at corresponding
3D locations.
Features are interpolated at local token centers.
Occupancy is resampled on the local voxel grid and max-pooled to the
same token resolution, yielding $\overline C_i$.
The resulting condition is
\begin{equation}
\begin{aligned}
\widetilde F_i(\mathbf u)
&=\operatorname{TriInterp}\!\left(F_i,T_i(\mathbf u)\right),\\
A_i(\mathbf u)
&=\left[
\operatorname{LN}\!\left(\widetilde F_i(\mathbf u)\right),
\overline C_i(\mathbf u)
\right].
\end{aligned}
\label{eq:aligned_condition}
\end{equation}
Each local token therefore receives both geometric support and
scene-conditioned features for the region it will generate.

The inherited frame also defines supervision for the subsequent
building generator.
We transform the ground-truth mesh $\mathcal M_i^{\star}$ by
$T_i^{-1}$ and voxelize it to obtain the local target $Y_i^{\star}$.
Generated local meshes are returned to the scene using $T_i$,
so conditioning, supervision, and assembly share the same coordinate
reference.

\subsection{Conditioned Building Regeneration}
\label{sec:local_generation}

To recover geometry beyond the coarse scaffold, we regenerate
local structure from noise, conditioned on the inherited
representation $A_i$ and image tokens $e_i$ extracted from
a tight building crop.

\paragraph{Spatially aligned conditioning.}
We adapt the pretrained sparse-structure flow model of
TRELLIS~\citep{xiang2025trellis} by injecting the inherited
condition after each transformer block,
\begin{equation}
\begin{aligned}
H_i^{\ell+1}(\mathbf u)
={}&\mathcal B_{\ell}(H_i^{\ell},t,e_i)(\mathbf u)\\
&+W_{\ell}A_i(\mathbf u)+b_{\ell},
\end{aligned}
\label{eq:injection}
\end{equation}
where $\mathcal B_{\ell}$ combines self-attention, image
cross-attention, and feed-forward operations.
Shared spatial indexing supplies each token with the inherited
occupancy and features of its corresponding local region.
The projections are zero-initialized to preserve the pretrained
computation at initialization.
Repeated injection keeps this spatial condition available
throughout the network and along the sampling trajectory.

\paragraph{Training objective.}
The frozen SS encoder maps $Y_i^{\star}$ to the target latent.
We train the conditional velocity field using the flow-matching
loss $\mathcal L_{\mathrm{FM}}$.
To directly supervise decoded geometry, we additionally recover
the estimated clean latent for low-noise samples and pass it
through the frozen SS decoder to obtain occupancy probabilities
$P_i$.
Our geometric objective combines soft Dice overlap with
centroid and axial-distribution constraints,
\begin{equation}
\begin{aligned}
\mathcal L_{\mathrm{geo},i}
={}&
\underbrace{
1-\frac{2\langle P_i,Y_i^{\star}\rangle+\delta}
{\|P_i\|_1+\|Y_i^{\star}\|_1+\delta}
}_{\mathcal L_{\mathrm{Dice},i}}\\
&+\lambda_c
\|\mu(P_i)-\mu(Y_i^{\star})\|_2^2\\
&+\operatorname{MAE}\!\left(m(P_i),m(Y_i^{\star})\right),
\end{aligned}
\label{eq:local_geometry}
\end{equation}
where $\delta$ is a smoothing constant, $\mu(X)$ denotes
the occupancy-weighted centroid in voxel-index coordinates,
and $m(X)$ concatenates the unit-mass occupancy marginals
along the three axes.
These terms constrain occupied support, centroid position,
and axial geometry distribution within the inherited frame.
The total objective is
$\mathcal L_{\mathrm{local}}
=\mathcal L_{\mathrm{FM}}+\lambda_g\mathcal L_{\mathrm{geo}}$,
where $\mathcal L_{\mathrm{geo}}$ is averaged over the selected
low-noise samples.
The flow formulation and hyperparameters are provided
in the appendix.

\paragraph{Mesh realization and scene assembly.}
We adopt the pretrained structured-latent (SLAT) generator
and mesh decoder of SAM~3D
Objects~\citep{sam3dteam2025sam3d}.
The regenerated sparse structure $\widehat Y_i$ defines
the spatial support for image-conditioned SLAT generation,
followed by mesh decoding.
After converting the resulting mesh to the inherited local
frame, we place it in the scene through
$\mathcal M_i=T_i(\widehat{\mathcal M}_i)$.
The assembled scene
$\mathcal S=\{\mathcal M_i\}_{i=1}^{N}$
retains individually editable building instances.

%% file: sections/03_dataset.tex
\section{CityLift-5K Dataset}
\label{sec:dataset}

To address the scarcity of urban scene data pairing images with complete building-level geometry, we construct CityLift-5K, comprising 4,997 synthetic urban blocks and 36,338 building instances. Each scene provides a rendered image, instance masks, and individually accessible building meshes in a shared coordinate frame, supporting supervision of both building geometry and spatial arrangement.

\paragraph{Asset preparation.}
We curate CityLift-Asset, a library of 42,339 building meshes from SAT-SKYLINES~\citep{jin2026satskylines}, CityCraft~\citep{deng2024citycraft}, UrbanVerse~\citep{liu2026urbanverse}, and subsets of TRELLIS-500K~\citep{xiang2025trellis,deitke2023objaversexl}.
We apply multiview screening with Qwen3-VL~\citep{bai2025qwen3vl}, followed by further manual filtering, to exclude non-building content, disconnected building groups, and assets with incomplete geometry.
The retained assets receive textual descriptions and annotations of function, architectural style, floor count, and estimated dimensions.
These attributes guide asset selection and provide scale priors for scene construction.

\paragraph{Scene construction and annotations.}
We combine MLLM-assisted layout planning with procedural parcel placement to assemble urban blocks on a planar ground.
The scenes cover residential, office, commercial, civic, and industrial settings, with 2--12 buildings per scene.
Buildings are selected according to their annotated function and size, then scaled and positioned within the planned layout.

We render each scene with Blender's Eevee engine, producing one $1024\times1024$ RGB image and a pixel-aligned instance segmentation map.
An orthographic camera captures an oblique view with an off-nadir angle sampled between $30^\circ$ and $40^\circ$.
Camera framing accounts for the block extent and maximum building height, while sun direction and intensity vary across scenes.
Each sample retains individual meshes, placement transforms, camera parameters, and instance identifiers.
The dataset contains 4,497 training, 250 validation, and 250 test scenes.
Further construction details and statistics are provided in the appendix.

%% file: sections/04_experiments.tex
\begin{table*}[!t]
\centering
\caption{\textbf{Geometric evaluation on CityLift-5K.} An em dash denotes unavailable building-level output.}
\label{tab:citylift-main}
\begingroup
\small
\setlength{\tabcolsep}{2.5pt}
\renewcommand{\arraystretch}{0.95}
\begin{tabular*}{\textwidth}{@{\extracolsep{\fill}}l*{11}{r}@{}}
\toprule
& \multicolumn{5}{c}{Building level}
& \multicolumn{5}{c}{Scene level}
& \multirow{2}{*}{\shortstack{Runtime$\downarrow$\\(s/scene)}} \\
\cmidrule(lr){2-6}\cmidrule(lr){7-11}
Method
& CD$\downarrow$ & Acc$\downarrow$ & Comp$\downarrow$
& F@1$\uparrow$ & F@2$\uparrow$
& CD$\downarrow$ & Acc$\downarrow$ & Comp$\downarrow$
& F@1$\uparrow$ & F@2$\uparrow$ & \\
\midrule
TRELLIS.2
& \textemdash & \textemdash & \textemdash & \textemdash & \textemdash
& 3.327 & 3.274 & 3.380 & 0.318 & 0.517
& \textbf{39.2} \\
SAM 3D
& 2.761 & 2.565 & 2.956 & 0.397 & 0.595
& 2.595 & 2.501 & 2.689 & 0.387 & 0.597
& 83.7 \\
Pixal3D
& 2.547 & 2.206 & 2.889 & 0.396 & 0.603
& 2.617 & 2.306 & 2.928 & 0.380 & 0.592
& 1482.4 \\
3D-Fixer
& 2.701 & 2.533 & 2.870 & 0.403 & 0.602
& 2.832 & 2.853 & 2.811 & 0.385 & 0.590
& 310.3 \\
SceneGen
& 9.352 & 10.092 & 8.612 & 0.126 & 0.223
& 4.973 & 4.795 & 5.151 & 0.214 & 0.368
& \multirow{2}{*}{231.9} \\
SceneGen-SFT
& 3.843 & 4.438 & 3.248 & 0.220 & 0.389
& 3.422 & 4.008 & 2.837 & 0.236 & 0.418
& \\
\midrule
\textbf{Ours}
& \textbf{1.011} & \textbf{1.065} & \textbf{0.956}
& \textbf{0.650} & \textbf{0.869}
& \textbf{1.079} & \textbf{1.138} & \textbf{1.021}
& \textbf{0.626} & \textbf{0.855}
& 115 \\
\bottomrule
\end{tabular*}
\endgroup
\end{table*}

\section{Experiments}
\label{sec:experiments}

We compare CityLift with representative baselines in terms of building-level geometric fidelity
and scene-level spatial consistency. Experiments cover the held-out
CityLift-5K test split and scene blocks constructed from UrbanScene3D and InstanceBuilding. We report geometric metrics
at both building and scene levels, complemented by a MLLM-as-a-judge visual evaluation.

\begin{figure*}[t]
    \centering
    \includegraphics[width=\textwidth]{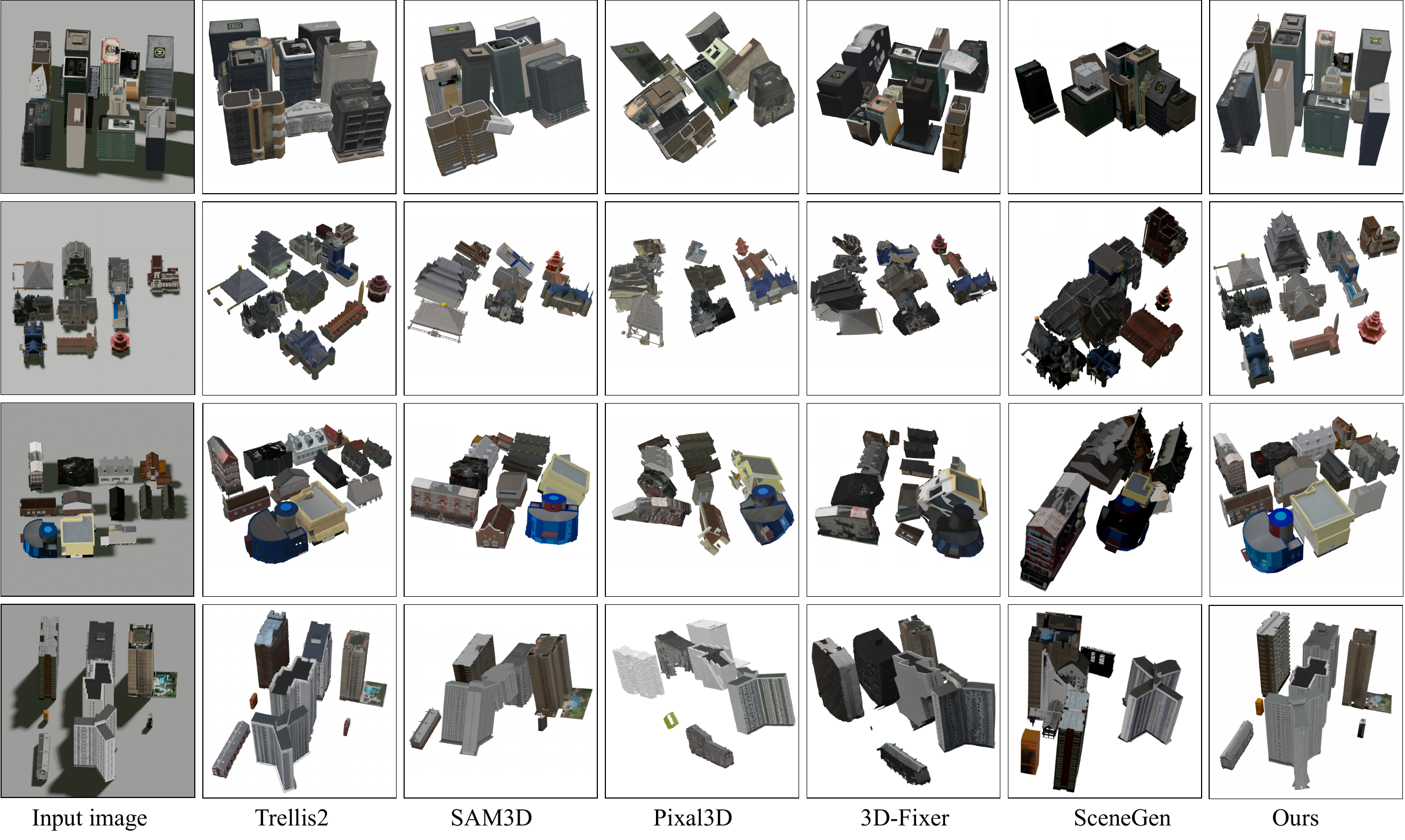}
    \caption{
        Qualitative comparison on blocks extracted from CityLift-5K testset. We slightly adjust the viewing angle to better showcase the generated scenes.
    }
    \label{fig:citylift-main-qualitative}
\end{figure*}

\subsection{Experimental Setup}
\label{sec:experiment-setup}

\paragraph{Datasets.}
The test split of CityLift-5K contains 250 scenes, averaging
7.8 buildings per scene. Both the scenes and their constituent
building assets are disjoint from the training split, ensuring
evaluation on unseen scenes and buildings. For external evaluation, we use the synthetic Shenzhen city
scene from UrbanScene3D~\cite{lin2022urbanscene3d}, extracting
its scene mesh and individual building meshes from Unreal
Engine. We also evaluate on Scene3 of
InstanceBuilding~\cite{chen20223D}, a real, densely built
town scene reconstructed from UAV imagery. Both scenes are partitioned into
blocks and rendered into RGB images and instance masks
using the CityLift data preparation pipeline. These tests
cover synthetic and photogrammetrically reconstructed
building geometry. 

\paragraph{Baselines.}
We compare with TRELLIS.2~\cite{xiang2025trellis2}, SAM 3D~\cite{sam3dteam2025sam3d}, Pixal3D~\cite{li2026pixal3d}, 3D-Fixer~\cite{yin2026fixer}, and SceneGen~\cite{meng2026scenegen}. TRELLIS.2 is applied to the entire scene image and produces a single scene mesh, so we report its scene-level metrics. The other baselines produce building instances that can be assembled into a scene. We additionally report SceneGen-SFT, fine-tuned on the CityLift training data, in the CityLift-5K comparison. This separates the released SceneGen model from its adaptation to the target domain.

\paragraph{Geometric evaluation.}
We evaluate geometry at both the building and scene levels using Chamfer distance (CD), accuracy (Acc), completeness (Comp), and F-scores at thresholds of one and two grid cells (F@1 and F@2). Lower distance errors and higher F-scores indicate better performance. Detailed evaluation protocols are provided in the appendix.

\paragraph{Implementation details.}
Both generation modules are trained on CityLift-5K using $64^3$
voxel grids. Training is performed on two NVIDIA A100 GPUs,
each with 80\,GB of memory. We use DINOv2~\cite{oquab2023dinov2} as the image feature extractor.
We first train the scaffold generator using AdamW with a
learning rate of $3.5\times10^{-5}$ and a global batch size
of 16. We then freeze the scaffold generator and train the
building regeneration module, initialized from
TRELLIS~\cite{xiang2025trellis}, for 18,000 iterations with
a learning rate of $5\times10^{-5}$.
At inference, pretrained SAM 3D converts the generated sparse
structures into textured meshes.
Further implementation details are provided in the appendix.

\subsection{Comparison on CityLift-5K}
\label{sec:citylift-main-results}
CityLift achieves the best geometric results at both building
and scene levels (Table~\ref{tab:citylift-main}). Building-level
CD decreases from 2.547 for Pixal3D to 1.011 cells, a reduction
of 60.3\%. Scene-level CD decreases from 2.595 for SAM 3D to
1.079 cells, a reduction of 58.4\%, while scene F@2 improves
from 0.597 to 0.855. CityLift requires approximately
$115$\,s per scene, faster than SceneGen, 3D-Fixer, and Pixal3D.

Figure~\ref{fig:citylift-main-qualitative} reveals differences
in building proportions and spatial arrangement.
TRELLIS.2 produces incorrect building heights, altering the
relative height relationships within a block. SAM 3D generates
recognizable individual buildings but introduces errors in
their relative positions and orientations. Pixal3D exhibits
flattened or fragmented geometry, while 3D-Fixer shows structural
distortions and less coherent spatial arrangements. SceneGen
produces pronounced scale inconsistencies and building overlaps.
CityLift more faithfully preserves building proportions,
relative orientations, and spacing, while retaining coherent
roof structures and facade details.

To assess whether dataset-specific fine-tuning can close the performance gap, we additionally fine-tune SceneGen on the CityLift-5K training split. SceneGen-SFT still yields higher building- and scene-level CDs of 3.843 and 3.422, compared with 1.011 and 1.079 for CityLift. These results suggest that adapting SceneGen to the same training data is insufficient to match CityLift's performance.

% \begin{figure*}[t]
% \centering
% \CLfigure{figures/citylift5k_comparison.pdf}{52mm}{CityLift-5K qualitative comparison}{Input RGB and masks; baseline outputs; CityLift; reference.\\Include a second view and enlarged building crops.}
% \caption{Qualitative comparison on CityLift-5K. Each example pairs the input with generated scenes and the reference geometry. Additional views and building crops expose differences that are difficult to inspect from the conditioning view alone.}
% \label{fig:citylift-main-qualitative}
% \end{figure*}

\subsection{Evaluation on External Scene Sources}
\label{sec:citylift-external}

\begin{table*}[t]
\centering
\caption{\textbf{External evaluation on UrbanScene3D.}
Quantitative comparison on blocks extracted from a synthetic
urban scene, assessing the geometric accuracy of individual buildings and assembled scenes.}
\label{tab:citylift-shenzhen}
\begingroup
\small
\setlength{\tabcolsep}{3.25pt}
\renewcommand{\arraystretch}{0.95}
\begin{tabular*}{\textwidth}{@{\extracolsep{\fill}}l*{10}{r}@{}}
\toprule
& \multicolumn{5}{c}{Building level} & \multicolumn{5}{c}{Scene level} \\
\cmidrule(lr){2-6}\cmidrule(lr){7-11}
Method & CD$\downarrow$ & Acc$\downarrow$ & Comp$\downarrow$ & F@1$\uparrow$ & F@2$\uparrow$
       & CD$\downarrow$ & Acc$\downarrow$ & Comp$\downarrow$ & F@1$\uparrow$ & F@2$\uparrow$ \\
\midrule
TRELLIS.2 & \textemdash & \textemdash & \textemdash & \textemdash & \textemdash & 2.809 & 2.766 & 2.851 & 0.401 & 0.616 \\
SAM 3D & 1.789 & 1.794 & 1.785 & 0.576 & 0.776 & 2.112 & 2.112 & 2.111 & 0.466 & 0.685 \\
Pixal3D & 1.490 & \textbf{1.303} & 1.678 & \textbf{0.600} & \textbf{0.818} & 2.067 & 1.666 & 2.469 & 0.473 & 0.693 \\
3D-Fixer & 1.797 & 1.497 & 2.098 & 0.489 & 0.708 & 2.507 & 1.973 & 3.042 & 0.422 & 0.641 \\
SceneGen & 13.376 & 13.634 & 13.118 & 0.083 & 0.158 & 8.017 & 5.566 & 10.467 & 0.147 & 0.248 \\
\midrule
\textbf{Ours} & \textbf{1.421} & 1.715 & \textbf{1.128} & 0.504 & 0.745 & \textbf{1.577} & \textbf{1.555} & \textbf{1.598} & \textbf{0.496} & \textbf{0.750} \\
\bottomrule
\end{tabular*}
\endgroup
\end{table*}

\begin{table*}[!t]
\centering
\caption{\textbf{External evaluation on InstanceBuilding.}
Quantitative comparison on scene blocks from real-world
reconstructed meshes, assessing the geometric accuracy of
individual buildings and assembled scenes.}
\label{tab:citylift-instancebuilding}
\begingroup
\small
\setlength{\tabcolsep}{3.25pt}
\renewcommand{\arraystretch}{0.95}
\begin{tabular*}{\textwidth}{@{\extracolsep{\fill}}l*{10}{r}@{}}
\toprule
& \multicolumn{5}{c}{Building level} & \multicolumn{5}{c}{Scene level} \\
\cmidrule(lr){2-6}\cmidrule(lr){7-11}
Method & CD$\downarrow$ & Acc$\downarrow$ & Comp$\downarrow$ & F@1$\uparrow$ & F@2$\uparrow$
       & CD$\downarrow$ & Acc$\downarrow$ & Comp$\downarrow$ & F@1$\uparrow$ & F@2$\uparrow$ \\
\midrule
TRELLIS.2 & \textemdash & \textemdash & \textemdash & \textemdash & \textemdash & 3.667 & 3.324 & 4.009 & 0.200 & 0.385 \\
SAM 3D & 1.995 & 2.421 & 1.570 & 0.366 & 0.620 & 2.418 & 2.887 & 1.949 & 0.351 & 0.594 \\
Pixal3D & 1.768 & \textbf{1.720} & 1.816 & \textbf{0.428} & \textbf{0.711} & 2.012 & 2.116 & 1.909 & 0.422 & 0.694 \\
3D-Fixer & 1.983 & 2.186 & 1.780 & 0.369 & 0.637 & 2.192 & 2.403 & 1.982 & 0.387 & 0.648 \\
SceneGen & 10.321 & 10.835 & 9.807 & 0.098 & 0.195 & 5.801 & 6.034 & 5.567 & 0.178 & 0.325 \\
\midrule
\textbf{Ours} & \textbf{1.625} & 1.915 & \textbf{1.334} & 0.400 & 0.684 & \textbf{1.531} & \textbf{1.790} & \textbf{1.272} & \textbf{0.428} & \textbf{0.717} \\
\bottomrule
\end{tabular*}
\endgroup
\end{table*}

\paragraph{UrbanScene3D.}
CityLift obtains the lowest building CD and leads all five scene-level metrics in Table~\ref{tab:citylift-shenzhen}. Its scene CD is 1.577 cells, compared with 2.067 for Pixal3D, and its scene F@2 is 0.750 compared with 0.693. These correspond to a 23.7\% reduction in scene CD and a 5.7 percentage-point increase in F@2. The gain persists when the input arrangement comes from an extracted urban scene rather than the procedural layouts used to construct CityLift-5K.

At the building level, Pixal3D achieves the lowest accuracy
error and the highest F-scores, indicating closer
prediction-to-reference surface agreement on average and
stronger correspondence within the evaluated thresholds.
This advantage is consistent with its pixel-aligned
conditioning, which back-projects image features into 3D
to establish explicit image--geometry
correspondence~\citep{li2026pixal3d}.
CityLift achieves a lower completeness error of 1.128 cells,
compared with 1.678 for Pixal3D, yielding a lower symmetric
CD of 1.421 versus 1.490 cells.
Since CD averages the two directional errors, the reduction
in completeness error outweighs the increase in accuracy
error. F-scores instead summarize surface precision and
recall within fixed distance thresholds.
CityLift thus provides better average reference-surface
coverage, while Pixal3D retains an advantage in thresholded
surface agreement for individual buildings.

\begin{figure*}[!t]
    \centering
    \includegraphics[width=\textwidth]{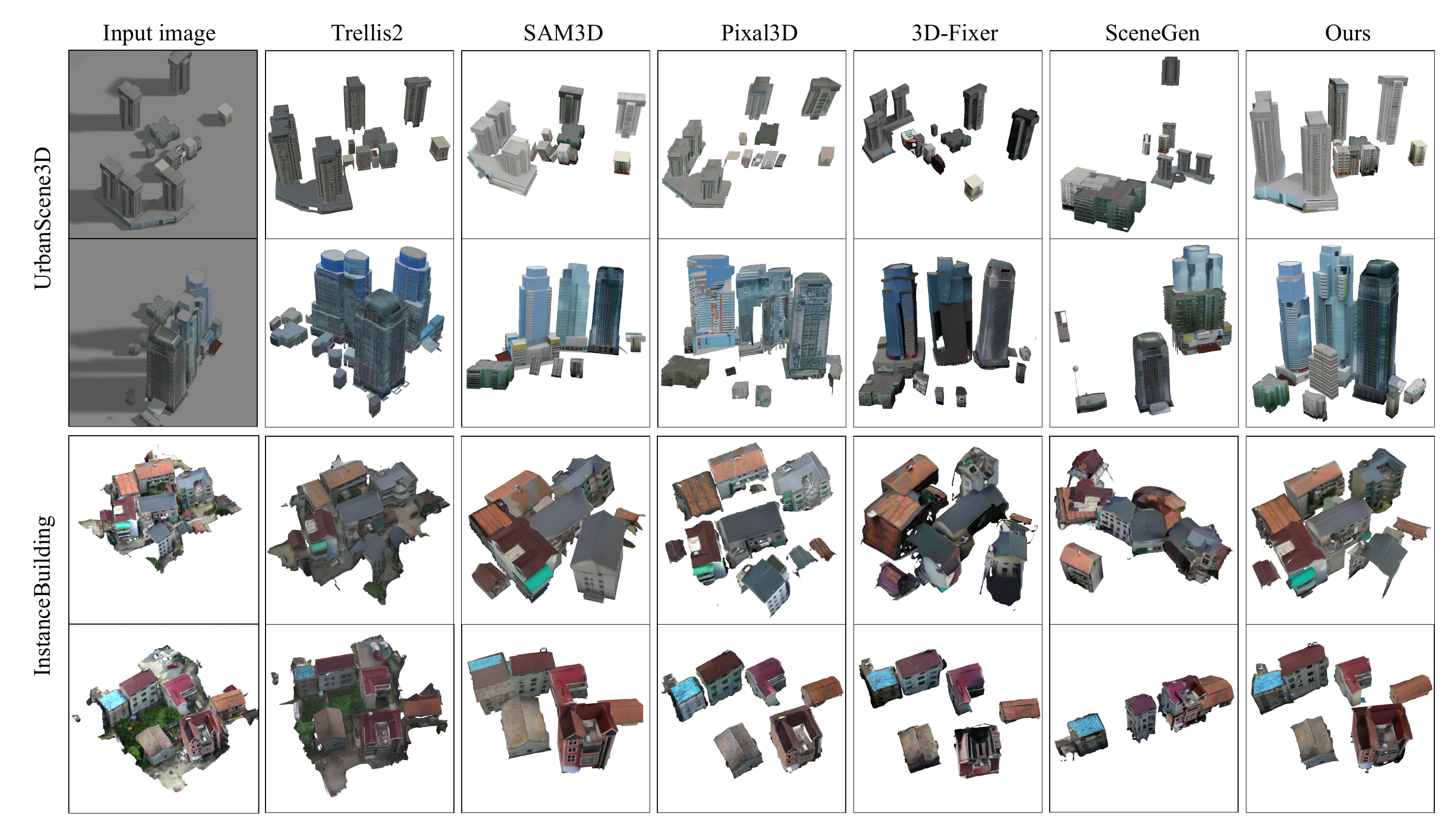}
    \caption{
        Qualitative comparison on blocks extracted from UrbanScene3D and InstanceBuilding. We slightly adjust the viewing angle to better showcase the generated scenes.
    }
    \label{fig:citylift-qualitative}
\end{figure*}

% \paragraph{InstanceBuilding.}
% CityLift again obtains the best values across all scene-level metrics, with a CD of 1.531 and F@2 of 0.717 (Table~\ref{tab:citylift-instancebuilding}). Relative to Pixal3D, the strongest baseline at the scene level, CD decreases by 23.9\% and F@2 increases by 2.3 percentage points. At the building level, CityLift achieves the lowest CD and completeness error, whereas Pixal3D retains the best accuracy and F-scores. This repeats the pattern observed on UrbanScene3D and supports the transfer of CityLift to the two evaluated scene sources. Figures~\ref{fig:citylift-qualitative} presents the corresponding visual comparisons.
\paragraph{InstanceBuilding.}
CityLift also leads all five scene-level metrics on
real-world scenes from InstanceBuilding
(Table~\ref{tab:citylift-instancebuilding}).
Compared with Pixal3D, scene CD decreases from 2.012 to
1.531 cells, a reduction of 23.9\%, and F@2 increases
from 0.694 to 0.717, an improvement of 2.3 percentage points.
The building-level results exhibit the same pattern.
Pixal3D retains the best accuracy and F-scores, whereas
CityLift reduces completeness error from 1.816 to 1.334
cells and CD from 1.768 to 1.625 cells.
Across both external datasets, CityLift consistently
improves average building-surface coverage and assembled-scene
geometry, demonstrating generalization to both synthetic
and real-world scene sources.
Figure~\ref{fig:citylift-qualitative} presents the corresponding
visual comparisons.

\subsection{Multi-view Visual Evaluation}
\label{sec:citylift-visual}

Surface distances describe geometric agreement with the reference, but do not measure texture quality or the visual quality of completed facades. We therefore render four views of each generated UrbanScene3D block and use ChatGPT 6 Astra as a visual assessor. For each method and block, the assessor inspects the conditioning image together with all four rendered views. It assigns scores to layout and completeness ($L$), shape and proportion ($G$), appearance and detail ($A$), and multi-view structural quality ($M$). The overall score is
\begin{equation}
S=10\bigl(0.35L+0.30G+0.20A+0.15M\bigr).
\label{eq:citylift-visual-score}
\end{equation}
 As shown in Table~\ref{tab:citylift-visual}, CityLift achieves the highest scores across all four dimensions, with an overall score of 82.8 compared with 77.6 for TRELLIS.2.

\begin{table}[t]
\centering
\caption{\textbf{MLLM-as-a-judge visual evaluation on UrbanScene3D.} ChatGPT 6 Astra scores the input image and four rendered views of each output. Results are averaged over all blocks shared by all six methods. All values are on a 0--100 scale, with higher values better. L, G, A, and M denote layout and completeness, shape and proportion, appearance and detail, and multi-view structural quality.}
\label{tab:citylift-visual}
\begingroup
\small
\setlength{\tabcolsep}{5pt}
\renewcommand{\arraystretch}{0.95}
\begin{tabular*}{\linewidth}{@{\extracolsep{\fill}}l*{5}{r}@{}}
\toprule
Method & L & G & A & M & Overall \\
\midrule
TRELLIS.2 & 84.2 & 75.2 & 71.7 & 74.8 & 77.6 \\
SAM 3D & 80.8 & 67.9 & 69.2 & 68.3 & 72.8 \\
Pixal3D & 79.6 & 64.2 & 63.8 & 64.6 & 69.5 \\
3D-Fixer & 80.8 & 65.8 & 58.5 & 65.6 & 69.6 \\
SceneGen & 44.0 & 55.8 & 53.1 & 54.8 & 51.0 \\
\midrule
\textbf{Ours} & \textbf{87.3} & \textbf{81.2} & \textbf{77.9} & \textbf{81.7} & \textbf{82.8} \\
\bottomrule
\end{tabular*}
\endgroup
\end{table}

\subsection{Ablation Studies}
\label{sec:citylift-ablations}

\begin{figure}[t]
    \centering
    \includegraphics[width=0.85\columnwidth]{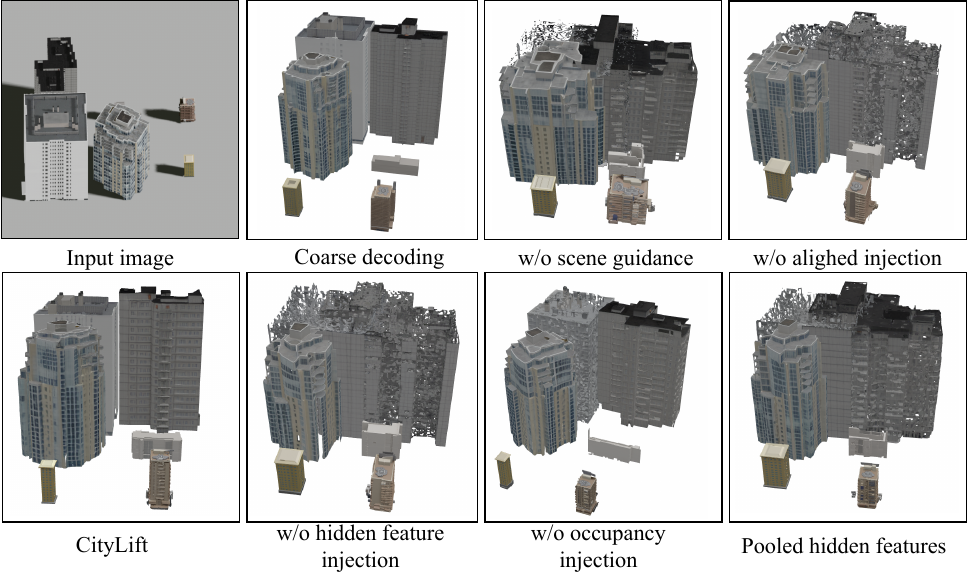}
    \caption{Qualitative results of the ablation study.}
    \label{fig:citylift-ablation-qualitative}
\end{figure}

We ablate scene guidance and replace local regeneration with
direct coarse decoding (Table~\ref{tab:citylift-ablations}).
Removing scene guidance causes the largest degradation,
increasing scene-level CD from 1.079 to 2.590 and highlighting
the importance of scene conditioning. Direct coarse decoding
achieves a building-level CD of 1.150, but produces smoother,
simplified surfaces with less distinct windows and other
facade details (Fig.~\ref{fig:citylift-ablation-qualitative}).
Local regeneration reduces CD to 1.011 and recovers finer
architectural details, complementing the overall building
geometry captured by the scaffold.

We further examine inherited conditioning by removing
hidden-feature and occupancy injection separately and jointly,
and replacing aligned hidden features with spatially pooled features.
Removing both injection pathways increases building-level CD
to 2.510, exceeding that of coarse decoding. Thus, local
regeneration can degrade an initially useful scaffold without
aligned conditioning. Removing hidden-feature injection and
pooling hidden features yield similar scene-level CDs of 2.170
and 2.200, suggesting that their benefit depends strongly on
preserving spatial variation. Occupancy injection provides a
smaller additional gain, with its removal increasing scene-level
CD to 1.260. Consistent with these findings,
Fig.~\ref{fig:citylift-ablation-qualitative} shows fragmented
building surfaces when aligned or hidden-feature injection is
removed, or hidden features are pooled, whereas CityLift
produces more complete facades and roofs.

\begin{table}[t]
\centering
\caption{Quantitative results of the ablation study on CityLift-5K.}
\label{tab:citylift-ablations}
\begingroup
\footnotesize
\setlength{\tabcolsep}{2.5pt}
\renewcommand{\arraystretch}{1.0}
\begin{tabular*}{\columnwidth}{@{\extracolsep{\fill}}lrrrr@{}}
\toprule
& \multicolumn{2}{c}{Building} & \multicolumn{2}{c}{Scene} \\[-1.5pt]
\cmidrule(lr){2-3}\cmidrule(lr){4-5}
Method & CD$\downarrow$ & F@2$\uparrow$
       & CD$\downarrow$ & F@2$\uparrow$ \\[-1.5pt]
\midrule
Coarse decoding & 1.15 & 0.85 & 1.22 & 0.84 \\
w/o scene guidance & 3.22 & 0.35 & 2.59 & 0.46 \\
w/o aligned injection & 2.51 & 0.47 & 2.41 & 0.50 \\
w/o hidden-feature injection & 2.21 & 0.56 & 2.17 & 0.58 \\
w/o occupancy injection & 1.18 & 0.85 & 1.26 & 0.83 \\
Pooled hidden features & 2.24 & 0.57 & 2.20 & 0.57 \\
\midrule
\textbf{Ours} & \textbf{1.011} & \textbf{0.869} & \textbf{1.079} & \textbf{0.855} \\
\bottomrule
\end{tabular*}
\endgroup
\end{table}

% \begin{figure}[t]
%     \centering
%     \includegraphics[width=\columnwidth]{images/citylift_ablation.pdf}
%     \caption{
%         Qualitative results of the ablation study.
%     }
%     \label{fig:citylift-ablation-qualitative}
% \end{figure}

%% file: sections/05_conclusion.tex
\FloatBarrier
\section{Conclusion}
\label{sec:conclusion}

We presented CityLift, a framework for compositional 3D urban generation from a single image and building instance masks. Urban scaffold generation captures scene layout and coarse building geometry, while generative spatial inheritance transfers geometric support and hidden features into local building frames. Conditioned building regeneration refines individual structures, which are converted into explicit, individually editable meshes and assembled using the inherited spatial transforms. We also introduced CityLift-5K to support training and evaluation of this task. Experiments on CityLift-5K, UrbanScene3D, and InstanceBuilding demonstrate improved scene-level geometric fidelity. These results support the effectiveness of combining scene-level spatial organization with local building generation for compositional urban content creation.